%% file: main.tex
\documentclass{article}
\usepackage{spconf,amsmath,graphicx}

\usepackage[colorlinks=true, linkcolor=blue, citecolor=red, urlcolor=magenta]{hyperref}

\usepackage{cite}
\usepackage{amssymb,amsfonts}
\usepackage{algorithmic}

\usepackage{textcomp}
\usepackage{xcolor}
\usepackage{booktabs}
\usepackage{makecell}
\usepackage{threeparttable}
\usepackage{subcaption}
\usepackage{colortbl,xcolor}

\usepackage[capitalize]{cleveref}
\crefname{section}{Sec.}{Secs.}
\Crefname{section}{Section}{Sections}
\Crefname{table}{Table}{Tables}
\crefname{table}{Tab.}{Tabs.}

\usepackage{multirow}
\usepackage{rotating}

\usepackage{arydshln}

\usepackage[most]{tcolorbox}

\newenvironment{remark}[1][]
  { 
 \begin{tcolorbox}
 [
    enhanced, 
    breakable,
    boxrule=0.5pt,
    arc=4pt,
    left=2pt,
    right=2pt,
    bottom=2pt,
    top=2pt, 
    rounded corners 
    ]{}
  \textbf{#1}
  \small \itshape
  }
  {
\end{tcolorbox} 
}

\usepackage{tikz}
\usepackage{graphicx}
\usepackage{xcolor}
\newcommand{\circleone}[1]{%
    \resizebox{!}{0.8em}{%
        \tikz[baseline=(char.base)]{
            \node[shape=circle, fill=black, inner sep=0.8pt, text=white] (char) {#1};
        }%
    }%
}

\newcommand{\circletwo}[1]{%
    \resizebox{!}{0.8em}{%
        \tikz[baseline=(char.base)]{
            \node[shape=circle, fill=black, inner sep=0.8pt, text=white] (char) {#1};
        }%
    }%
}

\title{Erosion Attack for Adversarial Training to Enhance Semantic Segmentation Robustness}
\name{Yufei Song$^{1}$, Ziqi Zhou$^{1}$, Menghao Deng$^{2}$, Yifan Hu$^{1}$, Shengshan Hu$^{1}$, Minghui Li$^{1}$, Leo Yu Zhang$^{3}$}

\address{
$^{1}$Huazhong University of Science and Technology\\
$^{2}$National University of Singapore\\
$^{3}$Griffith University\\
{\footnotesize\texttt{\{yufei17, zhouziqi, hyf1009, hushengshan, minghuili\}@hust.edu.cn}}\\
{\footnotesize\texttt{Dengmenghao17@u.nus.edu, leo.zhang@griffith.edu.au}}
}

\begin{document}
\maketitle
\input{Section/0-abstract}
\input{Section/1-introduction}

\input{Section/2-relatedwork}
\input{Section/3-methodology}

\input{Section/4-experiments}
\input{Section/5-conclusion}

{
\small
\bibliographystyle{IEEEbib}
\bibliography{strings,refs}
}

\end{document}

%% file: Section/0-abstract.tex
\begin{abstract}
Existing segmentation models exhibit significant vulnerability to adversarial attacks. 
To improve robustness, adversarial training incorporates adversarial examples into model training. 
However, existing attack methods consider only global semantic information and ignore contextual semantic relationships within the samples, limiting the effectiveness of adversarial training.
To address this issue, we propose EroSeg-AT, a vulnerability-aware adversarial training framework that leverages EroSeg to generate adversarial examples.
EroSeg first selects sensitive pixels based on pixel-level confidence and then progressively propagates perturbations to higher-confidence pixels, effectively disrupting the semantic consistency of the samples.
Experimental results show that, compared to existing methods, our approach significantly improves attack effectiveness and enhances model robustness under adversarial training.
\end{abstract}


%% file: Section/1-introduction.tex
\section{Introduction}
\label{sec:intro}
Semantic segmentation models power critical applications in autonomous driving~\cite{li2023mseg3d}, medical imaging~\cite{kalinin2020medical}, and remote sensing~\cite{macdonald2024vistaformer} by classifying pixels to delineate target regions. Despite their success, they remain highly vulnerable to adversarial examples~\cite{goodfellow2014explaining,madry2017towards,xie2017adversarial,zhou2024darksam, zhou2025sam2, wang2025advedm}, where small input changes can cause large output deviations and pose serious risks in practice.

\begin{figure}[t]
 \setlength{\abovecaptionskip}{4pt}
    \centering
    \includegraphics[scale=0.45]{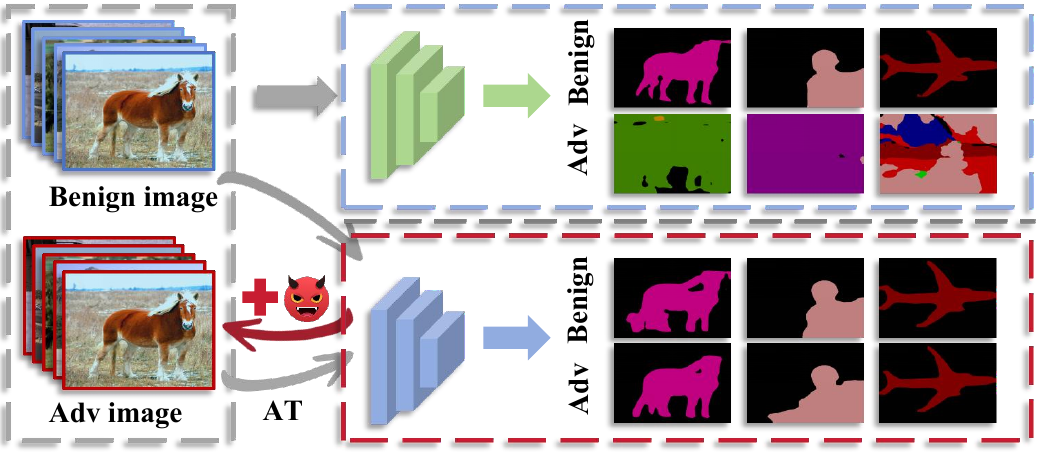}
    \caption{
    Illustration of adversarial training
    }
    \label{fig:demo}
  \vspace{-0.4cm}
\end{figure}

\textit{Adversarial training}~\cite{xu2021dynamic,gu2022segpgd,zhang2025rp} (AT) is widely used to improve the robustness of semantic segmentation models 
by training on adversarial examples generated during attacks, as illustrated in \cref{fig:demo}.
Although AT shows significant effectiveness in classification tasks, it remains underexplored in semantic segmentation.
Existing efforts, such as SegPGD~\cite{gu2022segpgd} and RP-PGD~\cite{zhang2025rp}, design tailored attacks to produce adversarial examples for training. However, they primarily optimize perturbations using global semantics, neglecting pixel-wise prediction difficulty and contextual dependencies, which segmentation models exploit to resist attacks.
In this work, we refer to this phenomenon as perturbation smoothing.

We argue that effective AT requires adversarial examples that fully expose model vulnerabilities.
The effectiveness of adversarial attacks on semantic segmentation hinges on both accurately targeting vulnerable pixels and exploiting contextual semantic relationships.
Our approach precisely identifies and attacks sensitive pixels that are most vulnerable to perturbations.
By altering the classification of these pixels, the attack progressively spreads to neighboring stable pixels.
Based on this, we propose EroSeg-AT, a vulnerability-aware adversarial training framework that leverages EroSeg to generate adversarial training samples. EroSeg consists of two modules:
\circleone{1} The first module, Confidence-based Selection of Sensitive Pixels, calculates the classification confidence of each pixel and uses an initial threshold to accurately identify and prioritize attacks on sensitive pixels. 
\circletwo{2}  The second module, Progressive Perturbation Propagation, employs a growth factor to dynamically increase the threshold during attack iterations, enabling the perturbation to progressively propagate.

Experiments on two mainstream segmentation models (DeepLabV3~\cite{xing2020encoder} and PSPNet~\cite{zhao2017pyramid}) and two benchmark datasets (PASCAL VOC~\cite{everingham2010pascal} and CITYSCAPES~\cite{cordts2016cityscapes}) show that EroSeg not only significantly outperforms existing methods in terms of attack strength but also, as a baseline attack for AT, notably improves the robustness of semantic segmentation models.
Our main contributions are as follows: 
(1) We reveal the relationship between pixel-level confidence in segmentation tasks and the network’s vulnerability.
(2) We investigate how the tight correlations of contextual semantic information in segmentation tasks affect adversarial attacks.
(3) Experimental results show that EroSeg achieves stronger attack performance across multiple models and datasets while effectively improving robustness in AT.

%% file: Section/2-relatedwork.tex
\section{Related Works}
\label{sec:relat}
\subsection{Semantic Segmentation Models}
Semantic segmentation models employ various architectures to capture rich image features and improve segmentation accuracy. Common components include Image Pyramid~\cite{wu2022fpanet}, Encoder-Decoder~\cite{xing2020encoder}, Context Modules~\cite{wu2020cgnet}, Spatial Pyramid Pooling~\cite{ru2023forest}, Atrous Convolution~\cite{chen2014semantic}, and Atrous Spatial Pyramid Pooling~\cite{chen2017deeplab}. These designs enhance the ability of models to extract multi-scale features, incorporate contextual information, and enlarge the receptive field, which together lead to more precise segmentation results.

\subsection{Adversarial Training}
Semantic segmentation models are shown to be vulnerable to adversarial attacks~\cite{ advclip, zhou2023downstream, zhou2025numbod, wang2025breaking,  song2025segment,  li2024transferable, song2025seg}. 
For this purpose, researchers propose various defense strategies, among which AT~\cite{zhou2024securely} stands out as one of the most effective methods for improving robustness due to its practicality and wide adoption. 
For example, the earliest DDCAT~\cite{xu2021dynamic} dynamically adjusts pixel branches based on boundary properties to enhance model robustness. 
SegPGD~\cite{gu2022segpgd} incorporates pixel-level weighting into AT to counter perturbations arising from differences in pixel importance. 
RP-PGD~\cite{zhang2025rp} further optimizes pixels in real regions and along target boundaries to provide stronger defense. 
These methods offer important ideas for improving the robustness of semantic segmentation models.

%% file: Section/3-methodology.tex
\section{Methodology}
\label{sec:methodology}

\noindent\textbf{Problem Formulation.}
Let $f_\theta(x)$ denote a segmentation model, where the input is $x_i$ with corresponding label $y_i$, and the output represents per-pixel class predictions. 
Let $\delta$ denote an adversarial perturbation applied to the input, constrained by $\|\delta\|_p \le \epsilon$, and let $\mathcal{L}$ denote the cross-entropy loss. 
Adversarial training improves model robustness by minimizing the loss on both clean and perturbed samples, where adversarial examples are generated under the $\ell_p$-norm constraint. 
Given $N$ training samples, the objective can be formulated as the following min-max optimization problem, where $w$ balances the losses on clean and adversarial samples:
\begin{equation}
\small
\min_{\theta}\; \frac{1}{N}\sum_{i=1}^N \Big(\mathcal{L}(f_\theta(x_i),y_i)
+ \lambda \max_{\|\delta_i\|_p\le\epsilon}\mathcal{L}(f_\theta(x_i+\delta_i),y_i)\Big),
\label{eq:adv_training}
\end{equation}

\begin{figure}[t]   
  \centering
      \subcaptionbox{Pixel choose}{\includegraphics[width=0.23\textwidth]{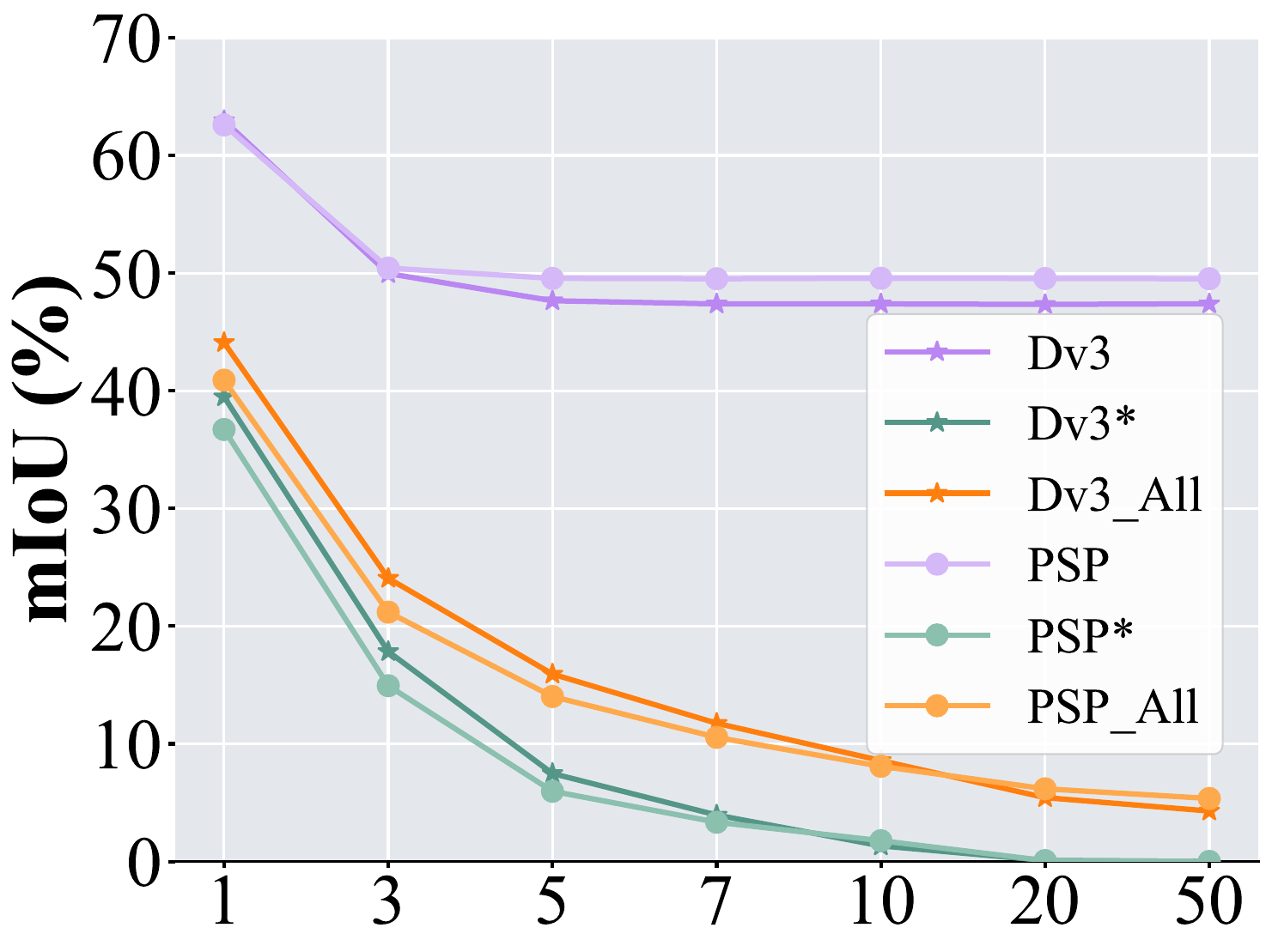}}
      \subcaptionbox{Target choose}{\includegraphics[width=0.23\textwidth]{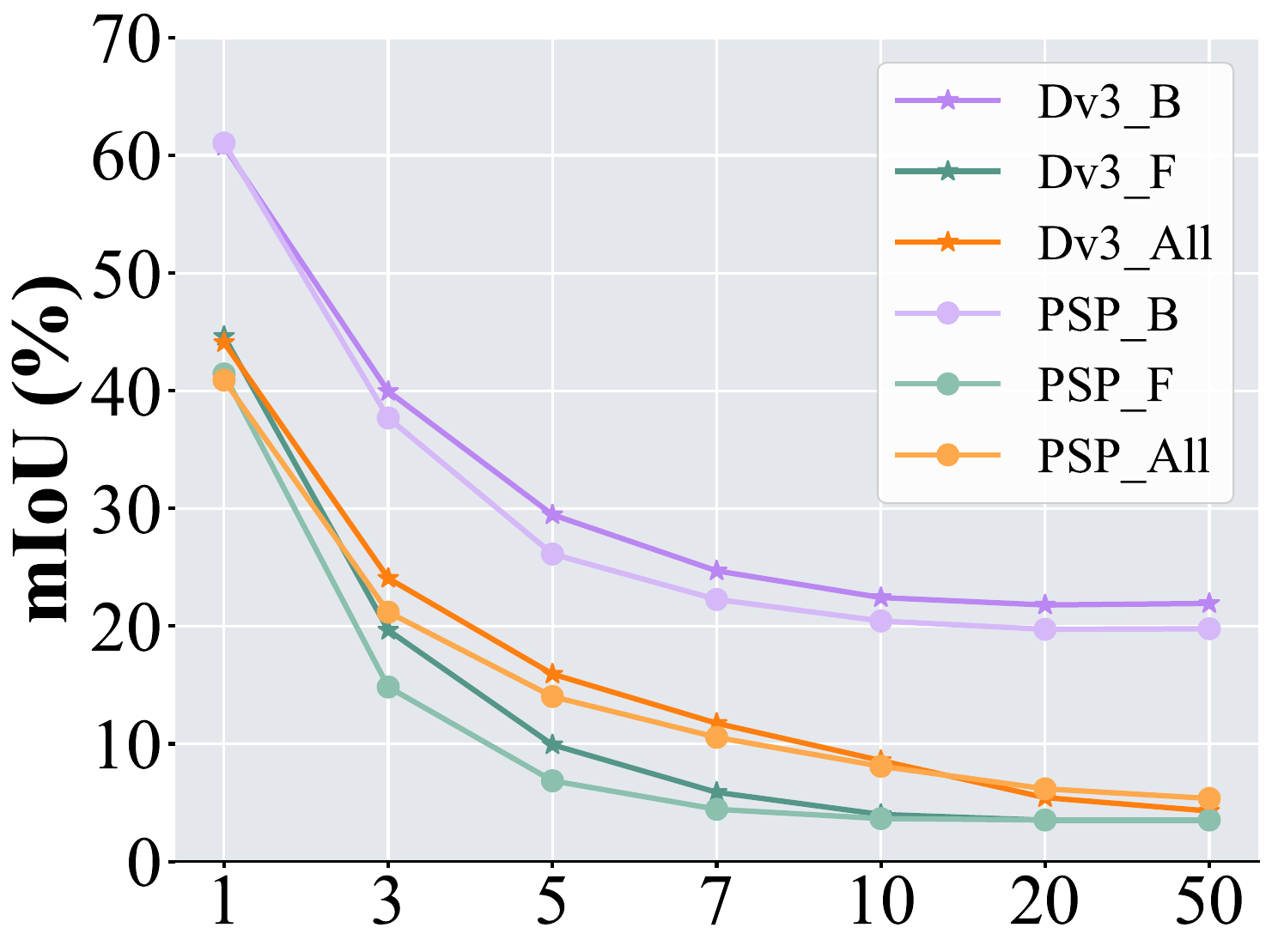}}
      \caption{
      Subfigure (a) shows the attack performance when targeting correctly classified pixels with confidence 1, those with confidence less than 1 (*), or all pixels (All). 
      Subfigure (b) shows the effect of attacking only background (B), foreground (F), or all pixels (All).
      }
       \label{fig:tanjiu}
        \vspace{-0.4cm}
\end{figure}

\vspace{-0.1cm}
\subsection{Motivation}
Unlike classification, segmentation attacks must disrupt both global semantics and local structural information, which makes them more challenging.
Specifically, we identify two key challenges for building baseline attacks for segmentation models.
\circleone{1} \textbf{Sensitive pixels are difficult to locate.}
Semantic segmentation requires the model to label every pixel, and pixels differ in prediction difficulty and vulnerability. 
Pixels near object boundaries or in regions with complex textures are more easily affected by perturbations, while other areas remain relatively stable. 
As shown in Fig.~\ref{fig:tanjiu} (a), although the attack targets correctly classified pixels, attacking pixels with a confidence of 1 has minimal effect.
\circletwo{2} \textbf{Perturbation smoothing caused by contextual dependencies.} 
Segmentation models rely on neighboring pixels and global semantic relationships to smooth or correct local predictions, which makes optimizing adversarial perturbations using only global semantic information insufficient to effectively disrupt model predictions.
As shown in Fig.~\ref{fig:tanjiu} (b), because foreground pixels contain critical target information and lie near semantic boundaries, adversarial examples optimized on the foreground achieve stronger attack performance than those optimized using global semantic information.

 \begin{figure*}[!t]
    \centering
    \includegraphics[scale=0.5]{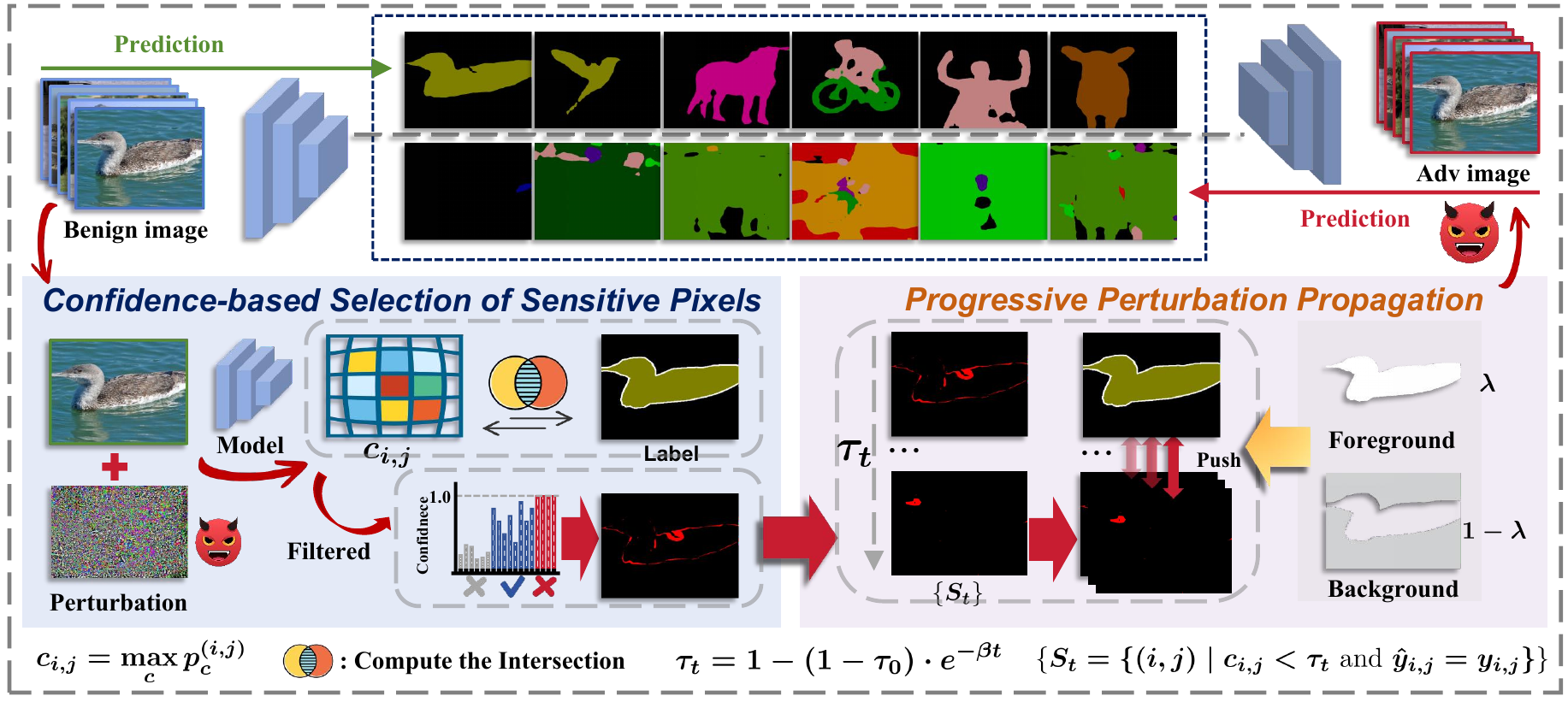}
    \caption{The framework of EroSeg}
    \label{fig:pipeline}
      \vspace{-0.4cm}
\end{figure*}

\begin{remark}[Takeaway.]
Existing methods determine pixel vulnerability solely by classification correctness, ignoring confidence differences, which hinders accurate identification of truly sensitive pixels. They also do not consider the smoothing effect of contextual semantic relationships, which reduces perturbation effectiveness.

\end{remark}

\subsection{EroSeg: A Complete Illustration}
In this section, we present EroSeg, a baseline adversarial attack method for AT of semantic segmentation models. 
As shown in~\cref{fig:pipeline}, EroSeg consists of two modules: (1) confidence-based selection of sensitive pixels and (2) progressive perturbation propagation.
In the first module, EroSeg measures pixel vulnerability using classification confidence and prioritizes pixels with lower confidence for targeted perturbations.
In the second module, we employ a progressive attack strategy that extends perturbations from sensitive pixels to more stable ones, enabling spatial propagation and overcoming the model’s contextual correction mechanisms. 
At the same time, the optimization assigns higher attention to foreground pixels, as shown in Fig.~\ref{fig:tanjiu} (b), to balance the uneven proportion of foreground and background pixels, thereby improving overall attack effectiveness.

\noindent\textbf{Confidence-based Selection of Sensitive Pixels.}
In semantic segmentation and other dense prediction tasks, the prediction difficulty and vulnerability vary significantly across pixels. 
Pixels near object boundaries, in regions with complex textures, or belonging to small foreground objects are generally more susceptible to perturbations, while pixels in large homogeneous background regions remain relatively stable. 
To effectively leverage these differences in adversarial attacks, EroSeg identifies sensitive pixels based on pixel-wise classification confidence.
Let the model output be $f(x+\delta)$, the true pixel labels be $y$, and the pixel-wise prediction probabilities be $p_c^{(i,j)}$, where $(i,j)$ denotes the pixel location and $c$ the class index. The maximum confidence for each pixel is defined as:

\begin{equation}
c_{i,j} = \max_c p_c^{(i,j)}.
\label{eq:1}
\end{equation}

Pixels with low confidence that are correctly classified form the sensitive pixel set:
\begin{equation}
S_0 = \{(i,j) \mid c_{i,j} < \tau_0 \ \text{and} \ \hat{y}_{i,j} = y_{i,j}\},
\label{eq:2}
\end{equation}
where $\tau_0$ is 0.8, and $\hat{y}_{i,j} = \arg\max_c p_c^{(i,j)}$ denotes the model's current prediction. 

\noindent\textbf{Progressive Perturbation Propagation.}
EroSeg employs a progressive perturbation propagation strategy that adjusts the threshold $\tau_t$ with a growth factor $\beta$, gradually extending perturbations from vulnerable pixels to more stable regions to disrupt contextual semantic consistency.
The $\tau_t$ is defined as:
\begin{equation}
 \tau_t = 1 - (1 - \tau_0) \cdot e^{-\beta t}, \quad t=1,2,\dots,T.
\label{eq:3}
\end{equation}

Additionally, EroSeg introduces class-aware weights to emphasize foreground pixels, defined as $w_{i,j} = \lambda$ if $y_{i,j} \neq 0$, and $w_{i,j} = 1-\lambda$ otherwise.
The weighted masked loss at iteration $t$ is defined as:  
\begin{equation}
L_t = \frac{1}{N \cdot H \cdot W} \sum_{i,j} m_{i,j}^t \cdot w_{i,j} \cdot 
\mathcal{L}_{\text{CE}}\!\left(f(x+\delta)_{i,j}, y_{i,j}\right).
\label{eq:4}
\end{equation}

%% file: Section/4-experiments.tex
\section{Experiments}
\label{sec:experiments}

\begin{table*}[htbp]
\setlength{\abovecaptionskip}{4pt}
  \centering
  \caption{Attack results of AT models under different settings, where the performance of the original model is shown in the Ori column, and the bold values indicate the mIoU of the most robust model under the same setting.}
  \scalebox{0.68}{
    \begin{tabular}{ccccccccccccccccc}
    \toprule
    \toprule
    \multirow{2}[4]{*}{Settings} & \multicolumn{8}{c}{DeepLabv3}                                 & \multicolumn{8}{c}{PSPNet} \\
    \cmidrule(lr){2-9}\cmidrule(lr){10-17}
        & \multicolumn{4}{c}{Pascal VOC} & \multicolumn{4}{c}{Cityscapes} & \multicolumn{4}{c}{Pascal VOC} & \multicolumn{4}{c}{Cityscapes} \\
\cmidrule(lr){2-5}\cmidrule(lr){6-9}\cmidrule(lr){10-13}\cmidrule(lr){14-17}   AT Strategy & Ori & CW    & PGD10 & PGD20 & Ori & CW    & PGD10 & PGD20 & Ori & CW    & PGD10 & PGD20 & Ori & CW    & PGD10 & PGD20 \\
\cmidrule{2-17}    Standard &   76.23    &   12.91    & 8.60  & 5.50  &    60.09   &   13.67    & 4.04  & 1.54  &   72.84    &   13.42    & 8.11  & 6.24  &    59.23   &    14.29   & 6.12  & 2.75  \\
\hdashline
    PGD3-AT & 71.81  & 50.19  & 32.96  & 21.05  & 52.64  & 40.13 & 28.47 & 19.19 & 65.57  & 36.33  & 24.65  & 15.42  & 46.85  & 35.99 & 22.92 & 14.13 \\
    SegPGD3-AT & 72.38  & 47.68  & 31.54  & 20.49  & 51.68  & 37.66 & 26.55 & 16.63 & 65.66  & 35.17  & 23.87  & 15.03  & 51.64  & 37.27 & 25.52 & 15.90 \\
    CosPGD3-AT & 71.13  & 50.71  & 32.92  & 20.38  & 52.53  & 39.05 & 27.42 & 17.28 & 66.15  & 35.87  & 24.29  & 14.20  & 51.45  & 38.67 & 26.54 & 15.72 \\
    RP-PGD3-AT & 72.37  & 51.80  & 33.28  & 21.06  & 53.17  & 40.13 & 29.66 & 19.46 & 66.05  & 37.31  & 25.68  & 17.64  & 52.14  & 38.95 & 26.55 & 15.93 \\
    EroSeg3-AT & \textbf{73.53} & \textbf{52.61} & \textbf{34.58} & \textbf{21.62} & \textbf{53.35} & \textbf{42.38} & \textbf{30.54} & \textbf{20.99} & \textbf{69.50 } & \textbf{38.78} & \textbf{26.24} & \textbf{18.54} & \textbf{52.73} & \textbf{39.71} & \textbf{27.87} & \textbf{16.42} \\
\hdashline
    PGD7-AT & 65.84  & 52.63  & 38.01  & 25.13  & 46.69  & 40.73 & 30.61 & 21.92 & 56.24  & 36.55  & 28.38  & 18.00  & 49.58  & 36.32 & 29.57 & 20.70 \\
    SegPGD7-AT & 67.94  & 52.26  & 36.86  & 24.55  & 45.04  & 39.03 & 28.9  & 20.01 & 57.47  & 37.46  & 27.74  & 17.07  & 44.25  & 37.34 & 24.57 & 15.98 \\
    CosPGD7-AT & 66.07  & 51.57  & 37.22  & 25.08  & 46.29  & 39.53 & 31.04 & 23.42 & 55.77  & 35.97  & 29.42  & 19.00  & 45.16  & 37.96 & 29.22 & 20.81 \\
    RP-PGD7-AT & 69.28  & 52.77  & 39.06  & 26.01  & 50.54  & 40.53 & 32.84 & 24.62 & 62.25  & 37.92  & 30.51  & 20.05  & 50.03  & 37.26 & 30.08 & 20.87 \\
    EroSeg7-AT & \textbf{71.49} & \textbf{54.62} & \textbf{39.88} & \textbf{26.59} & \textbf{50.94} & \textbf{43.28} & \textbf{33.06} & \textbf{25.41} & \textbf{63.28} & \textbf{39.05} & \textbf{32.67} & \textbf{21.87} & \textbf{51.58} & \textbf{38.10} & \textbf{31.55} & \textbf{21.39} \\
    \bottomrule
    \bottomrule
    \end{tabular}%
    }
  \label{tab:main}%
\end{table*}%
\begin{figure*}[!t]   
\setlength{\abovecaptionskip}{4pt}
  \centering
       \subcaptionbox{$\tau$}
       {\includegraphics[width=0.24\textwidth]{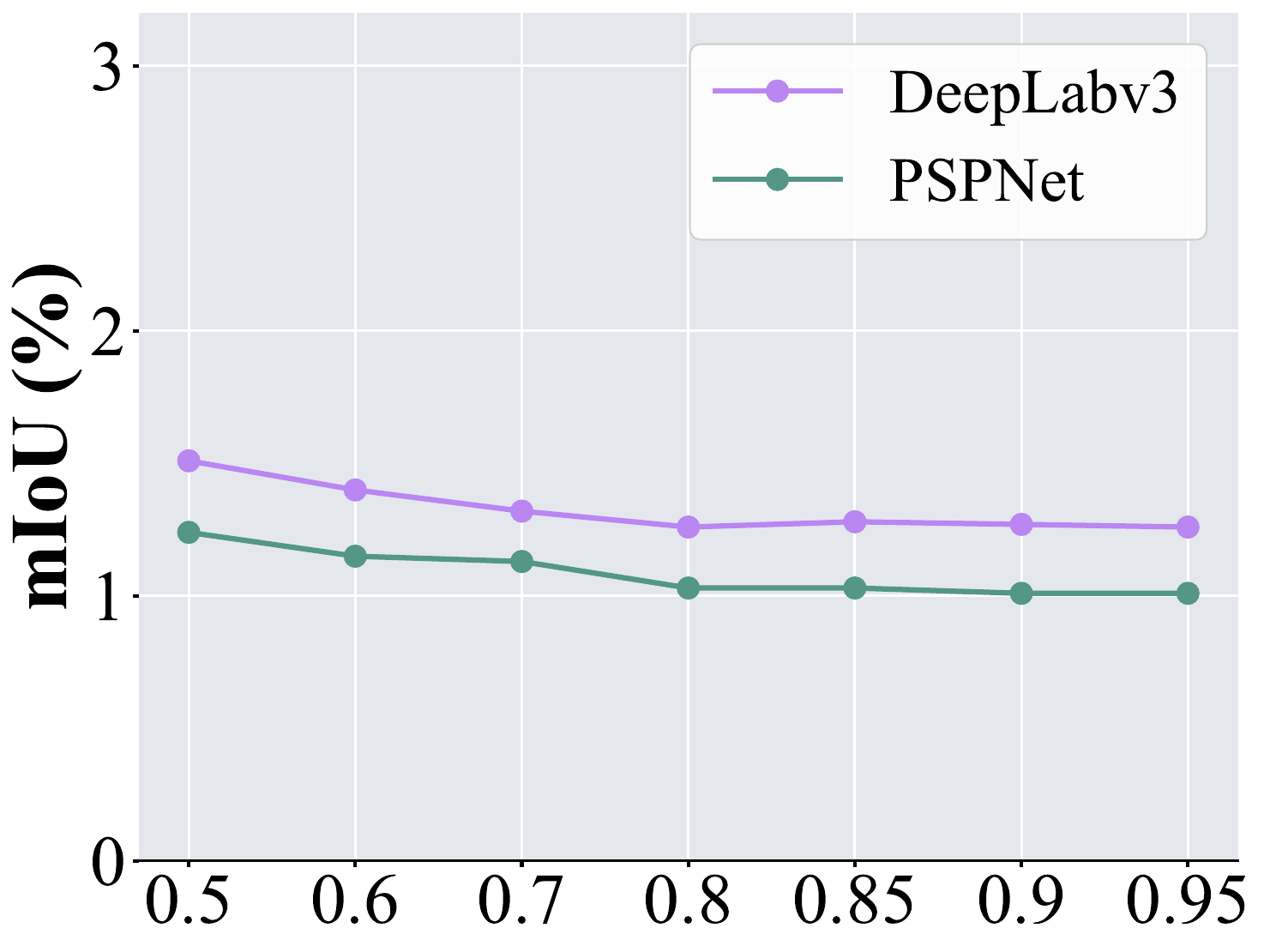}}
    \subcaptionbox{$\beta$}{\includegraphics[width=0.24\textwidth]{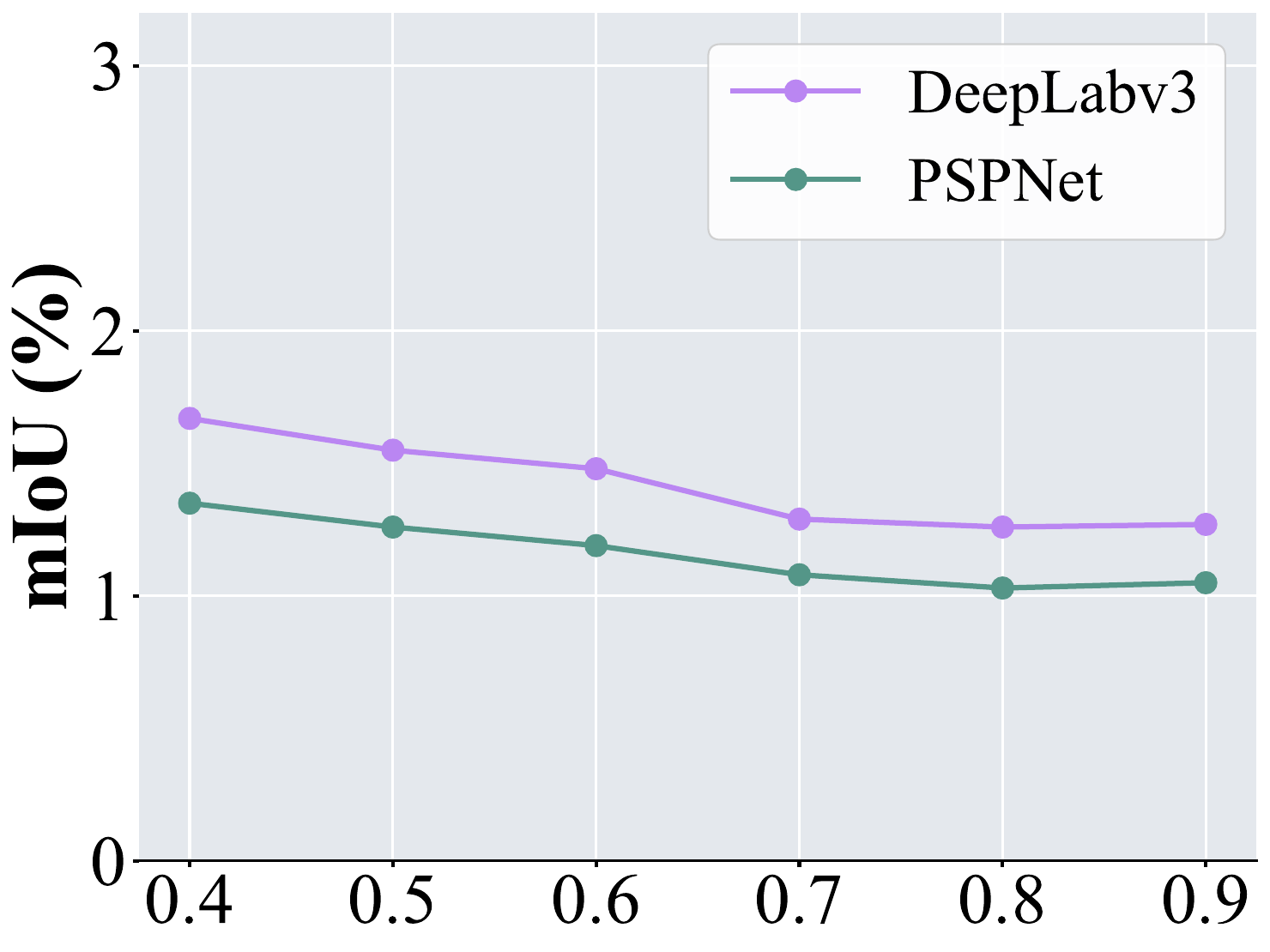}}
    \subcaptionbox{$\lambda$}{\includegraphics[width=0.24\textwidth]{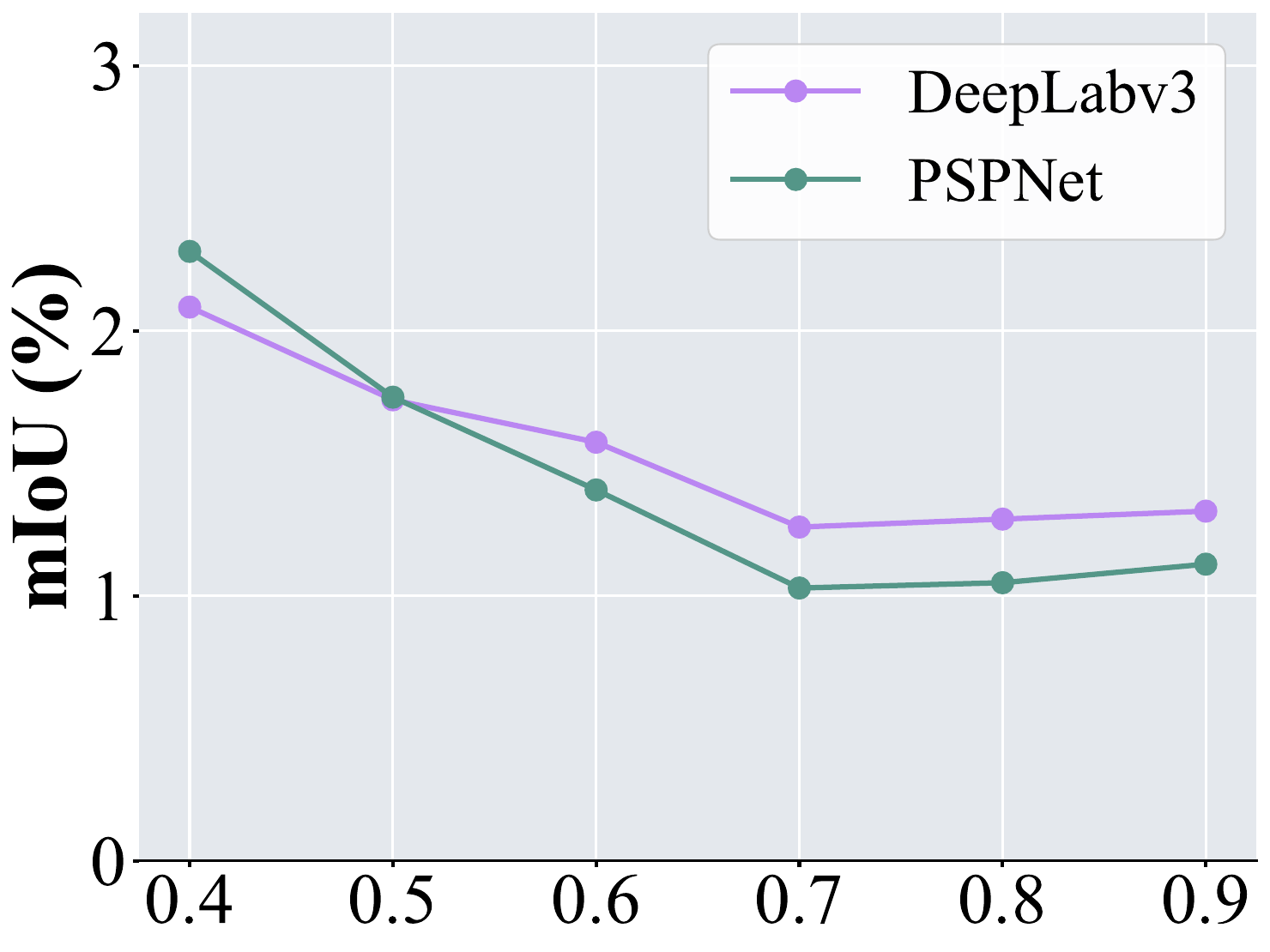}}
     \subcaptionbox{Iterations}{\includegraphics[width=0.24\textwidth]{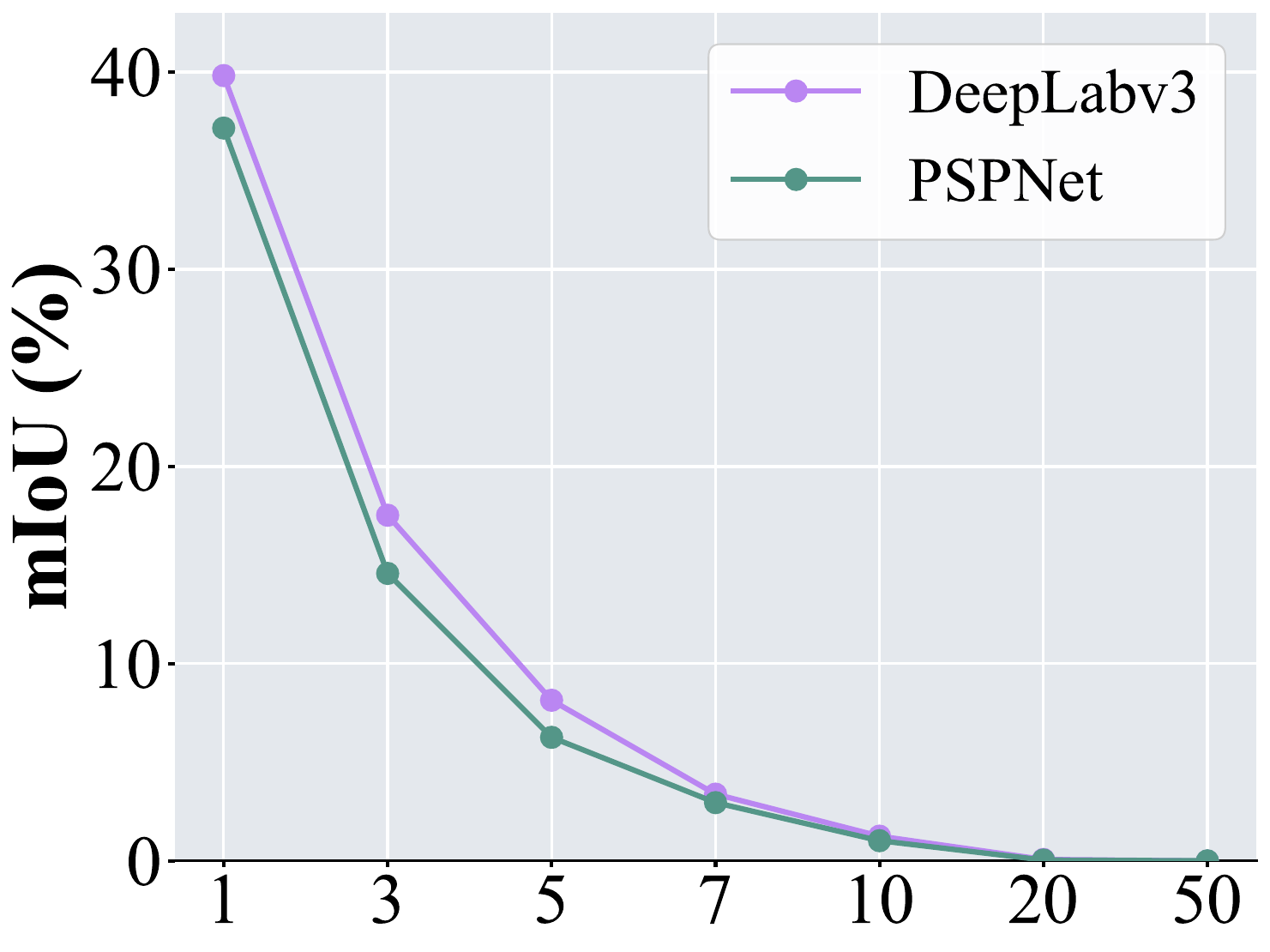}}
      \caption{The mIoU (\%) results of ablation study. 
      Including the effects of different initial thresholds (a), growth factors (b), weight factors (c), and attack iterations (d) on EroSeg.}
       \label{fig:ablation_results}
\end{figure*}

\begin{figure}[t]   
  \centering
      \subcaptionbox{DeepLabv3}{\includegraphics[width=0.23\textwidth]{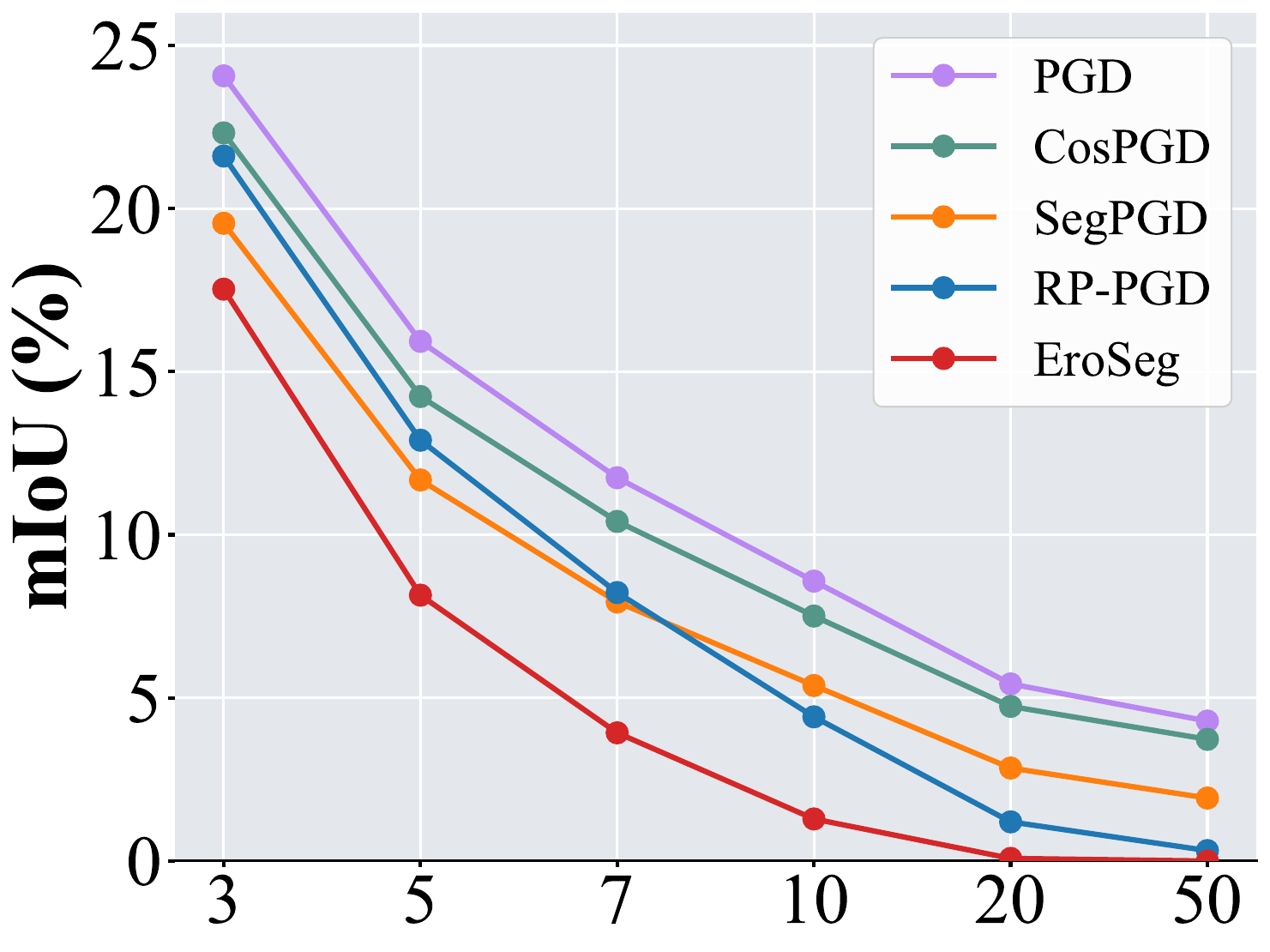}}
      \subcaptionbox{PSPNet}{\includegraphics[width=0.23\textwidth]{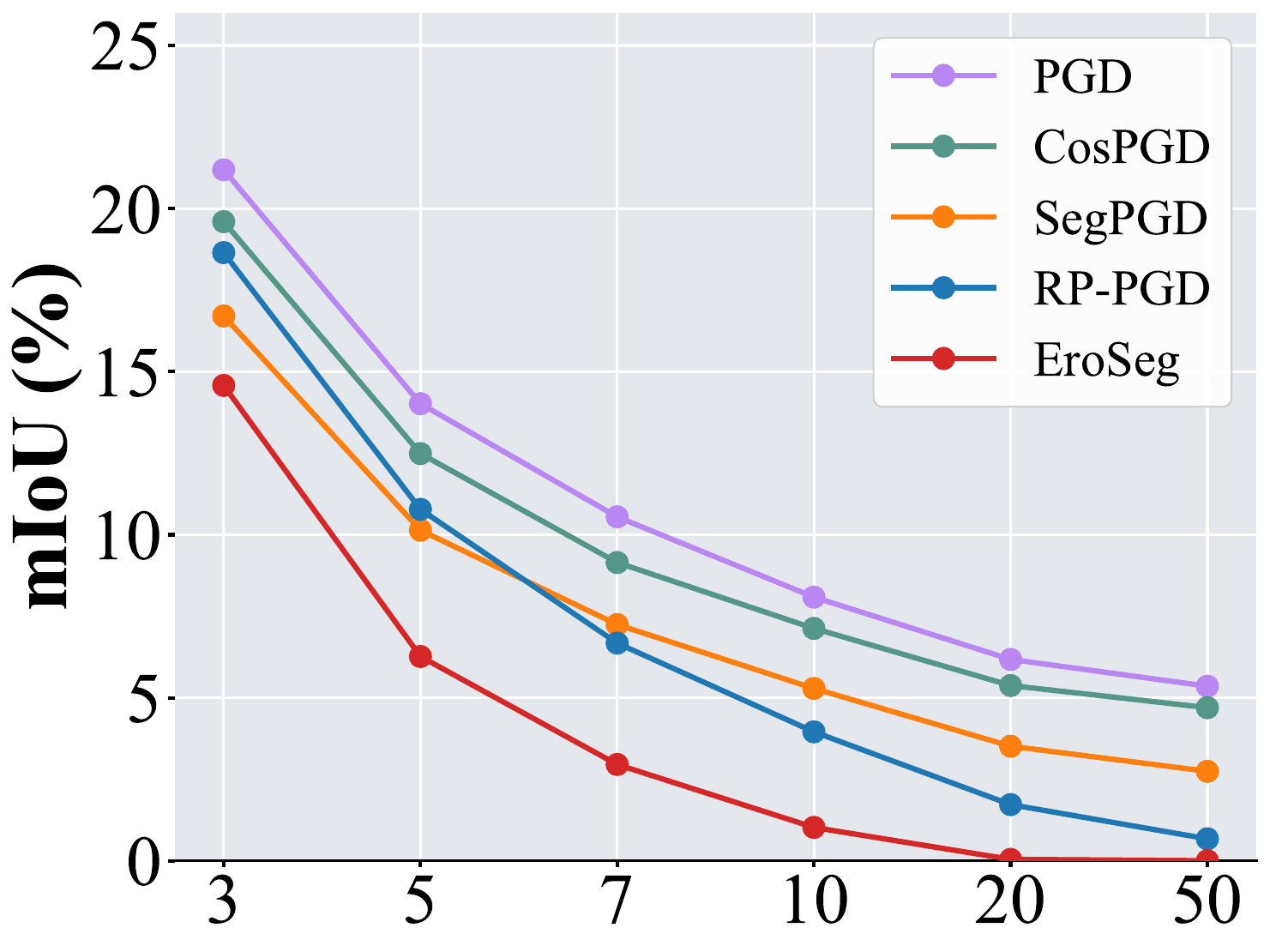}}
      \caption{The mIoU (\%) results of comparison study.}
       \label{fig:duibi}
        \vspace{-0.4cm}
\end{figure}

\subsection{Experimental Setup}
We evaluate the attack performance of EroSeg and its effectiveness in AT using two segmentation models, PSPNet~\cite{zhao2017pyramid} and DeepLabV3~\cite{xing2020encoder}, both with ResNet50 backbones. 
The evaluation is conducted on the PASCAL VOC~\cite{everingham2010pascal} and CITYSCAPES~\cite{cordts2016cityscapes} datasets. 
Following~\cite{gu2022segpgd,zhang2025rp,agnihotri2024cospgd}, we set the perturbation budget upper bound $\epsilon$ to $8/255$. 
The hyperparameters $\tau_0$, $\beta$, $\alpha$, and $\lambda$ are set to 0.8, 0.8, 0.001, and 0.7, respectively.
The evaluation metric used is \textit{mean Intersection over Union} (mIoU).

\vspace{-0.1cm}
\subsection{Attack Performance}
In this section, we evaluate the attack performance of EroSeg. 
Specifically, we conduct attacks on two segmentation models, DeepLab and PSPNet, using the PASCAL VOC dataset. 
We compare EroSeg with four mainstream segmentation attack methods, including PGD~\cite{madry2018towards}, CosPGD~\cite{agnihotri2024cospgd}, SegPGD~\cite{gu2022segpgd}, and RP-PGD~\cite{zhang2025rp}, as shown in Fig.~\ref{fig:duibi}. 
For each attack method, we compute the mIoU of the generated adversarial examples across 3 to 50 attack iterations. 
The results show that EroSeg consistently maintains a significant advantage across all iterations, and at the 10th iteration, the mIoU drops below 3\%, demonstrating the strong attack effectiveness of EroSeg.

\subsection{Ablation Study}
We analyze the effects of different hyperparameters on EroSeg using the PASCAL VOC dataset. 
The initial threshold $\tau$ ranges from $0.5$ to $0.95$, and the performance stabilizes at $\tau=0.8$, demonstrating that we achieve a good trade-off between attack success and computational efficiency (\cref{fig:ablation_results} (a)). 
The growth factor $\beta$ varies from $0.4$ to $0.9$, with stable performance at $\beta=0.8$ (\cref{fig:ablation_results} (b)). 
The weight $\lambda$ ranges from $0.4$ to $0.9$, reaching the best performance at $\lambda=0.7$ (\cref{fig:ablation_results} (c)). 
Finally, the number of iterations is set from $1$ to $50$, and when it reaches $7$, the mIoU drops below $5\%$, showing the strong attack capability of EroSeg (\cref{fig:ablation_results} (d)).

\vspace{-0.1cm}
\subsection{Adversarial training}
In this section, we evaluate the effectiveness of different adversarial attacks as benchmark attacks for adversarial training, including PGD, SegPGD, CosPGD, RP-PGD, and EroSeg. 
We use the subscript n to denote the number of iterations in an adversarial attack. 
For example, PGDn represents a PGD attack with n iterations, and PGDn-AT refers to AT using adversarial samples generated by PGD.
As shown in ~\cref{tab:main}, our method achieves the highest mIoU against all adversarial attacks. 
Notably, Our3-AT and Our7-AT reach average mIoUs of 30.86\% and 34.00\%, respectively, significantly higher than the existing methods, which achieve 28.65\% and 31.53\%. 
This clearly demonstrates the significant advantage of EroSeg as a baseline attack for AT of segmentation models.

%% file: Section/5-conclusion.tex
\section{Limitation and Conclusion}
\label{sec:conclusion}
In this work, we propose EroSeg, a strong baseline attack for AT of segmentation models. 
EroSeg identifies perturbation-sensitive pixels through classification confidence and applies a progressive attack strategy to disrupt contextual correction mechanisms, successfully deceiving the models. 
Extensive experiments show that, compared with existing methods, EroSeg achieves stronger attack performance and significantly improves adversarial robustness.
Moreover, our approach provides a new perspective for adversarial attacks and defenses in semantic segmentation, offering practical potential for advancing AT research.
A limitation of this work is that we have not yet extended EroSeg to detection tasks due to structural differences between segmentation and detection models. 
This remains a direction for future research.

\section*{Acknowledgements}
Shengshan Hu's work is supported by the National Natural Science Foundation of China under Grant No.62372196.
Minghui Li's work is supported by the National Natural Science Foundation of China under Grant No.62572206. 
Ziqi Zhou is the corresponding author.